\documentclass{article}

\usepackage{arxiv}

\usepackage[utf8]{inputenc} 
\usepackage[T1]{fontenc}    
\usepackage{url}            
\usepackage{booktabs}       
\usepackage{amsfonts}       
\usepackage{nicefrac}       
\usepackage{microtype}      

\usepackage{soul}
\usepackage{url}
\usepackage[hidelinks]{hyperref}
\usepackage[small]{caption}
\usepackage{graphicx}
\usepackage{amsmath,amsthm,amsfonts,amssymb}
\usepackage{subcaption}
\usepackage{booktabs}
\usepackage{algorithm}
\usepackage{algorithmic}
\usepackage[switch]{lineno}
\usepackage{xspace}
\usepackage{xcolor}

\usepackage{latexsym}

\newtheorem{theorem}{Theorem}
\newtheorem{lemma}{Lemma}
\newtheorem{proposition}{Proposition}
\newtheorem{corollary}{Corollary}
\theoremstyle{definition}
\newtheorem{definition}{Definition}
\newtheorem{example}{Example}

\newcommand{\atoms}{\ensuremath{\mathsf{At}}}
\newcommand{\langProp}[1]{\ensuremath{\mathcal{L}(#1)}\xspace}		
\newcommand{\allkbs}{\ensuremath{\mathbb{K}}}
\newcommand{\true}{\ensuremath{\mathsf{true}}}						
\newcommand{\false}{\ensuremath{\mathsf{false}}}						
\newcommand{\infer}{\ensuremath{\protect{|\hspace{-1.75mm}\sim}}}
\newcommand{\notinfer}{\ensuremath{\protect{\not\hspace{-1mm}\infer}}}
\newcommand{\infermaxcons}{\ensuremath{\infer^{\text{mc}}}}
\newcommand{\notinfermaxcons}{\ensuremath{\notinfer^{\text{mc}}}}
\newcommand{\varelem}{\ensuremath{\boxminus^{\text{ve}}}}
\newcommand{\naiveforget}{\ensuremath{\boxminus^{\text{na}}}}

\newcommand{\mc}{\ensuremath{\textsf{MCS}}}
\newcommand{\mi}{\ensuremath{\textsf{MIS}}}

\newcommand{\mcsig}{\ensuremath{\textsf{MCSig}}}
\newcommand{\misig}{\ensuremath{\textsf{MISig}}}
\newcommand{\scsig}{\ensuremath{\textsf{SCSig}}}

\newcommand{\mckb}{\ensuremath\textsf{MCKB}}

\newcommand{\interpretationsProp}[1]{\ensuremath{\Omega(#1)}}	
\newcommand{\modelSet}[1]{\ensuremath{\mathsf{Mod}(#1)}}			
\newcommand{\modelSetP}[1]{\ensuremath{\mathsf{Mod^3}(#1)}}			
\newcommand{\minModelSetP}[1]{\ensuremath{\mathsf{MinMod^3}(#1)}}			

\usepackage{environ}

\title{Reasoning with maximal consistent signatures}

\author{%
Matthias Thimm\\
Artificial Intelligence Group, University of Hagen, Germany\\
\texttt{matthias.thimm@fernuni-hagen.de}
\AND
Jandson S. Ribeiro\\
School of Computer Science and Informatics, Cardiff University, United Kingdom\\
\texttt{santosribeirosantosj@cardiff.ac.uk}
}

\begin{document}
\maketitle

\begin{abstract}
	We analyse a specific instance of the general approach of reasoning based on \emph{forgetting} by Lang and Marquis \cite{DBLP:conf/kr/LangM02,Lang:2010}. More precisely, we discuss an approach for reasoning with inconsistent information using \emph{maximal consistent subsignatures}, where a maximal consistent subsignature is a maximal set of propositions such that \emph{forgetting} the remaining propositions restores consistency. We analyse maximal consistent subsignatures and the corresponding minimal inconsistent subsignatures in-depth and show, among others, that the \emph{hitting set duality} applies for them as well. We further analyse inference relations based on maximal consistent subsignatures wrt.\ rationality postulates from non-monotonic reasoning and computational complexity. We also consider the relationship of our approach with inconsistency measurement and paraconsistent reasoning.
\end{abstract}

\section{Introduction}
Reasoning with inconsistent information is a central issue for approaches to knowledge representation and reasoning \cite{Rescher:1970,Grant:1978,DBLP:journals/sLogica/BenferhatDP97,DBLP:conf/kr/LangM02,Lang:2010,Arieli:2011,Konieczny:2019,Liu:2023}. A classical, yet simple, approach to reasoning with inconsistency is reasoning with maximal consistent subsets \cite{Rescher:1970,Konieczny:2019}. Given a (possibly inconsistent) knowledge base $K$ consisting of (propositional) formulas, a \emph{maximal consistent subset} $K'$ is a set $K'\subseteq K$ that is consistent and every set $K''$ with $K'\subsetneq K''\subseteq K$ is inconsistent (we will give formal definitions in Section~\ref{sec:prelim}). We can then define a simple inconsistency-tolerant inference operator $\infermaxcons_i$ on possibly inconsistent knowledge bases $K$ via $K\infermaxcons_i \alpha$ iff $K'\models \alpha$ for all maximal consistent subsets $K'$ of $K$ (where $\models$ is classical entailment). This inference operator has many good properties, in particular it coincides with classical entailment in the case of consistency. However, reasoning with maximal consistent subsets is highly syntax-dependent. In particular, it holds $\{a, \neg a , c\}\infermaxcons_i c$ but $\{a, \neg a \wedge c\}\notinfermaxcons_i c$.

In this work, we define and analyse a new approach to reason with maximal consistency, but defined in terms of subsignatures rather than subsets of the knowledge base. More precisely, we define a maximal consistent subsignature as a maximal set of propositions, such that \emph{forgetting}\footnote{Note that we use a slightly non-standard interpretation of the term \emph{forgetting}, in particular, we use forgetting to restore consistency, see Section~\ref{sec:forget} for details.} \cite{Lin:1994,DBLP:conf/aaai/WangSS05} the remaining propositions from the knowledge base makes it consistent again. This approach is a specific instance of the more general framework of Lang and Marquis \cite{DBLP:conf/kr/LangM02,Lang:2010}, who discuss inconsistency-tolerant reasoning under a wide variety of different strategies to forget propositions. However, the approach we will discuss has not been analysed explicitly nor are any of our results subsumed by the results of Lang and Marquis. Nonetheless, by considering both the notion of maximal consistent subsignatures and its counterpart, the minimal inconsistent subsignatures, we obtain a technical framework that is quite similar to the framework of maximal consistent subsets and minimal inconsistent subsets, but also features some additional interesting properties. We show that the classical \emph{hitting set duality} \cite{Reiter:1987} carries over to maximal consistent subsignatures as well, i.\,e., one can obtain maximal consistent subsignatures by removing a minimal hitting set of all minimal inconsistent subsignature, and vice versa. Furthermore, we show that inference relations based on maximal consistent subsignatures behave quite well in terms of rationality postulates for non-monotonic reasoning approaches \cite{Kraus:1990}. In order to complement our analysis, we also investigate the computational complexity of various problems pertaining to our approach and make a preliminary study on the application of the concepts of maximal consistent subsignatures and minimal inconsistent subsignatures for the purpose of \emph{inconsistency measurement} \cite{Grant:1978,Thimm:2019d}. Finally, we show that using a specific approach to forgetting, reasoning based on forgetting is equivalent to a specific paraconsistent logic, namely Priest's 3-valued logic \cite{Priest:1979}.

To summarise, the contributions of this paper are as follows:
\begin{enumerate}
	\item We revisit the notion of \emph{forgetting} parts of the signature of a knowledge base for the purpose of inconsistency-tolerant reasoning and make some new observations~(Section~\ref{sec:forget}).
	\item We define \emph{minimal inconsistent} and \emph{maximal consistent} subsignatures and analyse their properties; in particular, we show that these structures also obey the \emph{hitting set duality} (Section~\ref{sec:mcsig}).
	\item We consider reasoning with maximal consistent subsignatures, define corresponding inference relations, and analyse their properties (Section~\ref{sec:maxcons}).
	\item We analyse the computational complexity of various reasoning tasks with maximal consistent subsignatures (Section~\ref{sec:complexity}).
	\item We define new inconsistency measures based on minimal inconsistent subsignatures and maximal consistent subsignatures and make some preliminary observations (Section~\ref{sec:im}).
	\item We show that reasoning with maximal consistent subsignatures based on a naive approach to forgetting is equivalent to reasoning with Priest's paraconsistent logic (Section~\ref{sec:priest}).
\end{enumerate}
We will discuss necessary preliminaries in Section~\ref{sec:prelim} and conclude in Section~\ref{sec:conc}.

\section{Preliminaries}\label{sec:prelim}
Let $\atoms$ be some fixed propositional signature, i.\,e., a (possibly infinite) set of propositions, and let $\langProp{\atoms}$ be the corresponding propositional language constructed using the usual connectives $\wedge$ (\emph{conjunction}), $\vee$ (\emph{disjunction}), $\rightarrow$ (\emph{implication}), and $\neg$ (\emph{negation}).
\begin{definition}
	A knowledge base $K$ is a finite set of formulas $K\subseteq\langProp{\atoms}$. Let $\allkbs$ be the set of all knowledge bases.
\end{definition}
If $\Phi$ is a formula or a set of formulas we write $\atoms(\Phi)$ to denote the set of propositions appearing in $\Phi$.
For a set $\Phi=\{\phi_{1},\ldots, \phi_{n}\}$ let $\bigwedge \Phi = \phi_{1}\wedge\ldots\wedge\phi_{n}$ and $\neg \Phi = \{\neg \phi\mid \phi\in \Phi\}$.

Semantics to a propositional language are given by
\emph{interpretations} where an \emph{interpretation} $\omega$ on
\atoms\ is a function $\omega:\atoms\rightarrow\{\true,\false\}$. Let
$\interpretationsProp{\atoms}$ denote the set of all interpretations
for \atoms\ (with the convention that $\omega(\top)=\true$ and $\omega(\perp)=\false$). An interpretation $\omega$ \emph{satisfies} (or is a
\emph{model} of) an atom $a\in\atoms$, denoted by $\omega\models a$,
if and only if $\omega(a)=\true$. The satisfaction relation $\models$
is extended to formulas in the usual way.
For $\Phi\subseteq\langProp{\atoms}$ we also define $\omega\models \Phi$ if and only if $\omega\models\phi$ for every $\phi\in\Phi$.

In the following, let $\Phi,\Phi_{1},\Phi_{2}$ be formulas or sets of formulas. Define the set of models $\modelSet{\Phi}=\{\omega\in \interpretationsProp{\atoms}\mid \omega\models \Phi\}$.
We write $\Phi_{1}\models \Phi_{2}$ if $\modelSet{\Phi_{1}}\subseteq\modelSet{\Phi_{2}}$.
$\Phi_{1},\Phi_{2}$ are \emph{equivalent}, denoted by $\Phi_{1}\equiv \Phi_{2}$, if and only if $\modelSet{\Phi_{1}}=\modelSet{\Phi_{2}}$. Two sets $\Phi_1,\Phi_2$ are \emph{element-wise equivalent}, denoted by $\Phi_1\equiv^e \Phi_2$ iff there are functions $\rho_1:\Phi_1\rightarrow \Phi_2$ with $\phi\equiv\rho_1(\phi)$ for all $\phi\in \Phi_1$ and $\rho_2:\Phi_2\rightarrow \Phi_1$ with $\phi\equiv\rho_2(\phi)$ for all $\phi\in \Phi_2$.
If $\modelSet{\Phi}=\emptyset$ we also write $\Phi\models \perp$ and say that $\Phi$ is \emph{inconsistent} (or \emph{unsatisfiable}).

For a knowledge base $K$, a set $K'\subseteq K$ is a \emph{minimal inconsistent subset} of $K$ iff $K'\models\perp$ and for all $K''\subsetneq K$, $K''\not\models\perp$. A set $K'\subseteq K$ is a \emph{maximal consistent subset} of $K$ iff $K'\not\models\perp$ and for all $K''$ with $K'\subsetneq K''\subseteq K$, $K''\models\perp$. Let $\mi(K)$ and $\mc(K)$ denote the set of all minimal inconsistent subsets and the set of all maximal consistent subsets, respectively. A classical approach for reasoning with inconsistent knowledge bases is via reasoning with maximal consistent subsets \cite{Rescher:1970}. For a knowledge base $K$ and a formula $\alpha$, define $K\infermaxcons_i \alpha$ iff $K'\models \alpha$ for all $K'\in\mc(K)$ (\emph{inevitable consequences}). For a knowledge base $K$ and a formula $\alpha$, define $K\infermaxcons_w \alpha$ iff $K'\models \alpha$ for some $K'\in\mc(K)$ (\emph{weak consequences}).

\section{Forgetting and Projecting}\label{sec:forget}
A \emph{forgetting operator} is an operator that removes a given set of propositions from a signature of the knowledge base. Its initial motivation \cite{Lin:1994} was to be able to remove irrelevant parts of a knowledge base, while \emph{retaining} previous inferences as much as possible. There exists certain properties that such an operator should satisfy \cite{Lin:1994,DBLP:conf/aaai/WangSS05} and it makes sense (sometimes) to identify \emph{forgetting} with the \emph{variable elimination operation} (see below). However, our aim in this paper is to use forgetting as an operation that is able to \emph{restore} consistency, i.\,e., by removing ``conflicting'' parts of the signature of the knowledge base, we wish to end up with a consistent knowledge base. In particular, restoring consistency will retract a lot of inferences, which is then not aligned with the initial motivation for forgetting from above. Therefore, we will be using the following very general definition of forgetting and consider a wide range of possible forgetting operators.
\begin{definition}\label{def:forget}
	A \emph{forgetting operator} $\boxminus$ is any function $\boxminus:2^{\langProp{\atoms}}\times 2^\atoms\rightarrow 2^{\langProp{\atoms}}$ with\footnote{Note that we use the infix notation for the forgetting operation. Furthermore, for ease of presentation, we omit curly brackets for singleton sets where appropriate, i.\,e., we write $K\boxminus a$ instead of $K\boxminus\{a\}$.}
	\begin{enumerate}
		\item $\atoms(K\boxminus S)\subseteq\atoms(K)$,
		\item $\atoms(K\boxminus S)\cap S=\emptyset$,
		\item $K\boxminus\emptyset=K$,
		\item $(K\boxminus a)\boxminus b \equiv^e (K\boxminus b)\boxminus a$ and
		\item if $K$ is consistent then $K\boxminus S$ is consistent
	\end{enumerate}
	for every knowledge base $K$, $a,b\in\atoms(K)$, and $S\subseteq \atoms(K)$.
\end{definition}
In other words, given a knowledge base $K$ and some set of propositions $S\subseteq \atoms(K)$ mentioned in $K$, the forgetting $K'=K\boxminus S$ is a knowledge base that does not mention any proposition in $S$ anymore (and no further propositions are added). Conditions 3--5 model some further requirements to avoid artificial operators. Condition~3 states that forgetting the empty set does not syntactically alter the knowledge base. Condition~4 ensures that the order of forgetting does not change the result (up to semantical equivalence on the formula level). Note that it easily follows by induction that
	\begin{align*}
		&(\ldots((K\boxminus a_1)\boxminus a_2)\ldots \boxminus a_n)	\\
		\equiv^e &(\ldots((K\boxminus a_{\pi(1)})\boxminus a_{\pi(2)})\ldots \boxminus a_{\pi(n)})
	\end{align*}
	for any permutation $\pi:\{1,\ldots,n\}\rightarrow\{1,\ldots,n\}$. Condition~5 states that forgetting something from a consistent knowledge base does not introduce inconsistency. Note that the above definition does not contain any requirements on preservation of inferences under forgetting, as usually required by forgetting operators (such as $K\models \alpha$ with $\atoms(\alpha)\cap S=\emptyset$ should require $K\boxminus S\models \alpha$). We avoid such requirements on purpose (apart from the simple cases defined in items 3--5 of Definition~\ref{def:forget}), since we will mainly be dealing with inconsistent knowledge bases, for which (classical) inference trivialises.
	
	The canonical forgetting operation is the \emph{variable elimination operation} \cite{Lin:1994}, which we will also consider here in a slightly different manner. For that, let $\phi[\psi\rightarrow \psi']$ denote the propositional formula that is obtained from $\phi$ by simultaneously replacing each occurrence of $\psi$ in $\phi$ by $\psi'$.
	\begin{definition}\label{def:varelem}
		For a formula $\phi$ and some $a\in\atoms(\phi)$ define
		\begin{align*}
			\phi\varelem a & = \phi[a\rightarrow \top] \vee \phi[a\rightarrow \bot]
		\end{align*}
		For $S\subseteq\atoms(\phi)$ with $S=\{a_1,\ldots,a_k\}$ define\footnote{The order of the propositions in $S$ is arbitrary. Due to Proposition~\ref{prop:varelem:forget} below, the order does not matter, up to semantical equivalence.}
		\begin{align*}
			\phi\varelem S & = (\ldots((\phi\varelem a_1)\varelem a_2)\varelem \ldots \varelem a_k)
		\end{align*}
		and $\phi\varelem\emptyset=\phi$.
		For a knowledge base $K=\{\phi_1,\ldots,\phi_n\}$ write
		\begin{align*}
			 &K\varelem S \\
			=&  \{\phi_1\varelem (S\cap \atoms(\phi_1)),\ldots, \phi_n\varelem (S\cap \atoms(\phi_n))\}
		\end{align*}
	\end{definition}
Note that the above definition of variable elimination on knowledge bases $K\varelem S$ is \emph{formula-wise}, i.\,e., any particular proposition $a\in S$ is eliminated separately in each formula (instead of first taking the conjunction of all formulas in $K$ and the eliminating the proposition on this conjunction, which is usually the standard definition). This allows this variant of variable elimination to restore consistency, since propositions in different formulas may be substituted differently. 
	\begin{proposition}\label{prop:varelem:forget}
		$\varelem$ is a forgetting operator.
		\begin{proof}
			We check the conditions of Definition~\ref{def:forget}:
			\begin{enumerate}
				\item $\atoms(K\varelem S)\subseteq\atoms(K)$: follows directly by definition (no new propositions are introduced).
				\item $\atoms(K\varelem S)\cap S=\emptyset$: since all occurrences of a proposition $a\in S$ are replaced by either $\top$ or $\perp$, no proposition of $S$ appears in $K\varelem S$ any more.
				\item $K\varelem\emptyset=K$: follows directly by definition.
				\item $(K\varelem a)\varelem b \equiv^e (K\varelem b)\varelem a$:  follows from commutativity and associativity of the disjunction operator as well as the facts that $(\phi\vee \psi)[\alpha\rightarrow\beta]=\phi[\alpha\rightarrow\beta]\vee \psi[\alpha\rightarrow\beta]$ and $\phi[\alpha\rightarrow\beta][\alpha'\rightarrow\beta']\equiv\phi[\alpha'\rightarrow\beta'][\alpha\rightarrow\beta]$ for all formulas $\phi,\psi,\alpha,\beta$.
				\item if $K$ is consistent then $K\varelem S$ is consistent: if $\omega:\atoms(K)\rightarrow \{\true,\false\}$ is a model of $K$, then, clearly, the interpretation $\omega':\atoms(K\varelem S)$ with $\omega'(a)=\omega(a)$ for all $a\in\atoms(K)\setminus S$ is a model of $K\varelem S$.\qedhere
		\end{enumerate}
		\end{proof}
	\end{proposition}
As mentioned above, our variant of variable elimination does not necessarily preserve unsatisfiability, i.\,e., eliminating a proposition from an inconsistent knowledge base may result in a consistent knowledge base.
	\begin{example}\label{ex:k1k2}
		Consider the knowledge base $K_1=\{a, \neg a \wedge c\}$ which is obviously inconsistent. However,
		\begin{align*}
			K_1' & = K_1\varelem \{a\} = \{\top\vee\perp, (\neg \top \wedge c) \vee (\neg \perp \wedge c)\} \equiv \{c\}
		\end{align*}
		is consistent. But for the inconsistent $K_2=\{a\wedge \neg a, c\}$ we get
		\begin{align*}
			K_2' & = K_2\varelem \{a\} = \{ (\top\wedge \neg \top) \vee (\perp\wedge \neg \perp), c\} \equiv \{\perp, c\}
		\end{align*}
		which is still inconsistent. Note that $\varelem$ cannot restore consistency in $K_2$ as the formula $a\wedge \neg a$ is in itself inconsistent.
	\end{example}
From now on, we will only consider knowledge bases, where each individual formula is consistent by itself (but there may be inconsistencies across different formulas).
	
We will also consider another variant of variable elimination, which will prove useful later (in particular in Section~\ref{sec:priest}, where we show that reasoning based on this operation coincides with reasoning with a particular paraconsistent logic). Instead of operating \emph{formula-wise} as above, this variant works \emph{occurrence-wise}. Let $\phi[\psi\rightarrow \psi_1'/\psi_2'/\ldots/\psi_n']$ denote the propositional formula that is obtained from $\phi$ by replacing the first occurrence of $\psi$ in $\phi$ by $\psi'_1$, the second occurrence of $\psi$ in $\phi$ by $\psi'_2$, and so on (the operation is undefined if the number of occurrences of $\psi$ in $\phi$ is not equal to $n$).
	\begin{example}
		For the formula $\phi=a\wedge (b\vee a) \wedge \neg a$ we have
		\begin{align*}
			\phi[a\rightarrow \top/\top/\perp] = \top \wedge (b\vee \top) \wedge \neg \perp \equiv b
		\end{align*}
	\end{example}
The above operation allows us to define a new variant of variable elimination as follows. Let $\#^\phi a$ denote the number of occurrences of $a\in\atoms(\phi)$ in $\phi$.
	\begin{definition}
		For a formula $\phi$ and some $a\in\atoms(\phi)$ define\footnote{The abbreviation ``na'' stands for \emph{naive} (variable elimination) operator.}
		\begin{align*}
			\phi\naiveforget a & = \bigvee_{x_1,\ldots,x_{\#^\phi a}\in\{\top,\perp\}}\phi[a\rightarrow x_1/\ldots/x_{\#^\phi a}]
		\end{align*}
	\end{definition}
The operator $\naiveforget$ is extended to sets of propositions $S\subseteq \atoms(\phi)$ and knowledge bases $K$ analogously as in Definition~\ref{def:varelem}. The operator $\naiveforget$ allows the replacement of each occurrence of $a$ with $\top$ or $\perp$ such that contradictions within a formula can be resolved. Although this might not be apparant from the syntactical definition, but what the operator, basically, does is the following: it replaces every disjunction containing $a$ or $\neg a$ by $\top$ and removing all occurrences of $a$ and $\neg a$ from conjunctions. From the perspective of inconsistency-tolerant reasoning, this allows for more flexibility, as we will see later. It should also be obvious that $\naiveforget$ satisfies our conditions on forgetting operators.
	\begin{proposition}
		$\naiveforget$ is a forgetting operator.
		\begin{proof}
			The proof is analogous to the proof of Proposition~\ref{prop:varelem:forget}.
		\end{proof}
	\end{proposition}
	\begin{example}\label{ex2:k1k2}
		We continue Example~\ref{ex:k1k2}. We get
		\begin{align*}
			K_1'' & = K_1\naiveforget \{a\} = \{\top\vee\perp, (\neg \top \wedge c) \vee (\neg \perp \wedge c)\} \equiv \{c\}
		\end{align*}
		which is the same as using $\varelem$. However, for $K_2$ we get
		\begin{align*}
			K_2' & = K_2\naiveforget \{a\}\\
			 &= \{ (\top\wedge \neg \top) \vee (\perp\wedge \neg \perp) \vee (\top\wedge \neg \perp) \vee (\perp\wedge \neg \top), c\}\\
			 & \equiv \{c\}
		\end{align*}
	\end{example}
The above observation can be formally phrased as follows.
	\begin{proposition}\label{prop:varelem:vs:naiveforget:1}
		Let $\phi$ be a formula such that $\phi\models\perp$.
		\begin{enumerate}
			\item For every $S\subseteq \atoms(\phi)$, $\phi\varelem S\models\perp$.
			\item There is $S\subseteq \atoms(\phi)$ such that $\phi\naiveforget S\not\models\perp$
		\end{enumerate}
		\begin{proof}
			Let $\phi$ be a formula such that $\phi\models\perp$.
			\begin{enumerate}
				\item Assume there is $S=\{a_1,\ldots,a_n\}\subseteq \atoms(K)$ such that $\phi\varelem S\not\models\perp$. Observe that
					\begin{align*}
						\phi\varelem S \equiv \bigvee_{x_1,\ldots,x_n\in\{\top,\perp\}}	\phi[a_1\rightarrow x_1]\ldots[a_n\rightarrow x_n]
					\end{align*}
					So $\phi\varelem S$ is consistent iff (at least) one of the disjuncts of the above representation is consistent. Let $\omega:\atoms(K)\setminus S\rightarrow\{\true,\false\}$ be a model of $\phi\varelem S$ of and let $\hat{x}_1,\ldots,\hat{x}_n\in\{\top,\perp\}$ be such that $\omega\models\phi[a_1\rightarrow x_1]\ldots[a_n\rightarrow x_n]$. Then define $\omega':\atoms(K)\rightarrow\{\true,\false\}$ via
						\begin{enumerate}
							\item $\omega'(a)=\omega(a)$ for all $a\in\atoms(K)\setminus S$,
							\item $\omega'(a_i)=\true$ for all $a_i\in S$ with $\hat{x}_i=\top$, and
							\item $\omega'(a_i)=\false$ for all $a_i\in S$ with $\hat{x}_i=\perp$.
						\end{enumerate}
					Then $\omega'\models \phi$, contradicting $\phi\models \perp$.
				\item This follows directly from observing that $\phi\naiveforget\atoms(\phi)\equiv \top$.\qedhere
			\end{enumerate}
		\end{proof}
	\end{proposition}
	The above proposition generalises to knowledge bases as follows.
	\begin{corollary}\label{cor:varelem:vs:naiveforget:1}
		Let $K$ be a knowledge base with $\phi\in K$ being a formula such that $\phi\models\perp$.
		\begin{enumerate}
			\item For every $S\subseteq \atoms(K)$, $K\varelem S\models\perp$.
			\item There is $S\subseteq \atoms(K)$ such that $K\naiveforget S\not\models\perp$
		\end{enumerate}
		\begin{proof}
			Let $K$ be a knowledge base with $\phi\in K$ being a formula such that $\phi\models\perp$.
			\begin{enumerate}
				\item Since $\phi\varelem (S\cap \atoms(K))\models \perp$ for every $S\subseteq \atoms(K)$ due to Proposition~\ref{prop:varelem:vs:naiveforget:1}, it necessarily follows $K\varelem S\models\perp$.
				\item This follows directly from observing that $K\naiveforget \atoms(K)\equiv \top$.\qedhere
			\end{enumerate}
		\end{proof}
	\end{corollary}
The above observations show that, from the perspective of inconsistency-tolerant reasoning, $\naiveforget$ may actually be a sensible choice for a forgetting operation, since it allows the restoration of consistency in any case.

Computationally-wise, however, the operator $\naiveforget$ can be characterised through $\varelem$ as follows.
\begin{proposition}\label{prop:varelem:vs:naiveforget:2}
	Let $\phi$ be a formula and $a\in\atoms(\phi)$ and let $a^1,\ldots,a^{\#^\phi a}$ be new atoms not appearing in $\phi$. Then
	\begin{align*}
		\phi\naiveforget a & = \phi[a\rightarrow a^1/\ldots/a^{\#^\phi a}]\varelem \{a^1,\ldots,a^{\#^\phi a}\}
	\end{align*}
	\begin{proof}
		Observe that
		\begin{align*}
			& \phi[a\rightarrow a^1/\ldots/a^{\#^\phi a}]\varelem \{a^1,\ldots,a^{\#^\phi a}\} \\
			=&  	\hspace*{-2mm}\bigvee_{x_1,\ldots,x_{\#^\phi a}\in\{\top,\perp\}}\hspace*{-8mm}\phi[a\rightarrow a^1/\ldots/a^{\#^\phi a}][a^1\rightarrow x_1]\\
			& \hspace*{4.7cm}\vspace*{-2cm}\ldots[a^{\#^\phi a}\rightarrow x_{\#^\phi a}]\\
			=& \hspace*{-2mm}\bigvee_{x_1,\ldots,x_{\#^\phi a}\in\{\top,\perp\}}\hspace*{-8mm}\phi[a\rightarrow x_1/\ldots/x_{\#^\phi a}]\\
			=&  \phi\naiveforget a\qedhere
		\end{align*}
	\end{proof}
\end{proposition}
In other words, $\phi\naiveforget a$ is the same as variable elimination after we replace each occurrence of $a$ with a new distinguished proposition first. This generalises to knowledge bases and sets of propositions as follows.
\begin{corollary}\label{cor:naive:varelem}
	Let $K$ be a knowledge base and $S\subseteq \atoms(\phi)$. Then
	\begin{align*}
		K\naiveforget S & = \{\phi[a_1\rightarrow a_1^1/\ldots/a_1^{\#^\phi a}]\ldots[a_n\rightarrow a_n^1/\ldots/a_n^{\#^\phi a}]\\
					& \qquad \varelem\{a_1^1,\ldots,a_1^{\#^\phi a},\ldots,a_n^1,\ldots,a_n^{\#^\phi a}\}\mid\\
			& \qquad  \phi\in K, \{a_1,\ldots,a_n\} = S\cap\atoms(K)\}
	\end{align*}
	\begin{proof}
		Follows directly from Proposition~\ref{prop:varelem:vs:naiveforget:2}.
	\end{proof}
\end{corollary}

The two forgetting operators \varelem\ and \naiveforget\ are only two possible choices of forgetting operators following Definition~\ref{def:forget}. For future work, we intend to investigate further operators, but for the remainder of this paper, we restrict ourselves to statements on those two operators and some general statements, where such are possible.

A forgetting operator $\boxminus$ allows us to project the signature of a knowledge base to a subset of its signature. We define this concept in a general manner as follows.
	\begin{definition}
		Let $\boxminus$ be a forgetting operator. For a knowledge base $K$ and $S\subseteq \atoms(K)$, the \emph{projection} of $K$ onto $S$ (wrt.\ $\boxminus$), denoted $K|^\boxminus_S$, as
		\begin{align*}
			K|^\boxminus_S	& = K\boxminus(\atoms(K)\setminus S)
		\end{align*}
	\end{definition}
	\begin{example}
		We continue Example~\ref{ex2:k1k2}. We get
		\begin{align*}
			K_1|^{\varelem}_{\{c\}}	& = \{c\} & K_2|^{\varelem}_{\{c\}}	& = \{\perp,c\}\\
			K_1|^{\naiveforget}_{\{c\}}	& = \{c\} & K_2|^{\naiveforget}_{\{c\}}	& = \{c\}
		\end{align*}
	\end{example}
\section{Minimal inconsistent and maximal consistent subsignatures}\label{sec:mcsig}
The notion of projection allows us to define analogues to the concepts of minimally inconsistent subsets and maximally consistent subsets of a knowledge base $K$, based on a more semantical perspective. In general, we say that a set $S\subseteq\atoms(K)$ is a \emph{consistent subsignature} (wrt.\ $\boxminus$) iff $K|^\boxminus_S$ is consistent, otherwise it is called an \emph{inconsistent subsignature}.
	\begin{definition}
		Let $K$ be a knowledge base and $\boxminus$ a forgetting operator.
		\begin{enumerate}
			\item $S\subseteq \atoms(K)$ is called a \emph{minimal inconsistent subsignature} of $K$ (wrt.\ $\boxminus$) if
				\begin{enumerate}
					\item $K|^\boxminus_S\models\perp$ and
					\item for all $S'$ with $S'\subsetneq S$, $K|^\boxminus_{S'}\not\models\perp$.
				\end{enumerate}
			\item $S\subseteq \atoms(K)$ is called a \emph{maximal consistent subsignature} of $K$ (wrt.\ $\boxminus$) if
				\begin{enumerate}
					\item $K|^\boxminus_S\not\models\perp$ and
					\item for all $S'$ with $S\subsetneq S'\subseteq \atoms(K)$, $K|^\boxminus_{S'}\models\perp$.
				\end{enumerate}
		\end{enumerate}
		Let $\misig^\boxminus(K)$ and $\mcsig^\boxminus(K)$ denote the set of all minimal inconsistent subsignatures (wrt.\ $\boxminus$) and the set of all maximal consistent subsignatures (wrt.\ $\boxminus$), respectively.
	\end{definition}
We furthermore say that a proposition $a\in\atoms(K)$ is a \emph{free proposition} in $K$ (wrt.\ $\boxminus$) iff $a\notin S$ for all $S\in\misig^\boxminus(K)$.
	\begin{example}
		We consider again the knowledge base $K_1=\{a,\neg a \wedge c\}$. Here we have
		\begin{align*}
			\misig^{\naiveforget}(K_1) &= \{\{a\}\}\\
			\mcsig^{\naiveforget}(K_1) &= \{\{c\}\}
		\end{align*}
		For $K_2=\{a\wedge \neg a, c\}$ we get likewise
		\begin{align*}
			\misig^{\naiveforget}(K_2) &= \{\{a\}\}\\
			\mcsig^{\naiveforget}(K_2) &= \{\{c\}\}
		\end{align*}
		For both cases, $c$ is also a free proposition.
	\end{example}
	\begin{example}\label{ex:mimc}
		Consider
		\begin{align*}
			K_3 & = \{a\wedge b \wedge d, \neg a \vee \neg b, b\wedge \neg c, (c \vee \neg b)\wedge d\}
		\end{align*}
		Here we get
		\begin{align*}
			\misig^{\naiveforget}(K_3) &= \{\{a,b\},\{b,c\}\}\\
			\mcsig^{\naiveforget}(K_3) &= \{\{a,c,d\},\{b,d\}\}
		\end{align*}
		and $d$ is a free proposition.
	\end{example}
Some straightforward observations are as follows.
	\begin{proposition}\label{prop:misc}
		Let $K$ be a knowledge base and $\boxminus$ a forgetting operator.
		\begin{enumerate}
			\item $K$ is consistent iff $\misig^\boxminus(K)= \emptyset$ iff $\mcsig^\boxminus(K)=\{\atoms(K)\}$.
			\item $\mcsig^{\varelem}(K)\neq \emptyset$ iff there is no $\phi\in K$ with $\phi\models\perp$.
			\item $\mcsig^{\naiveforget}(K)\neq \emptyset$.
		\end{enumerate}
		\begin{proof}
			Let $K$ be a knowledge base and $\boxminus$ a forgetting operator.
			\begin{enumerate}
				\item Let $K$ be consistent. Due to property 5 of Definition~\ref{def:forget}, for every $S\subseteq\atoms(K)$, $K|^{\boxminus}_S=K\boxminus \atoms(K)\setminus S$ is consistent as well. It follows $\misig^\boxminus=\emptyset$. From $\misig^\boxminus=\emptyset$ it follows that $\atoms(K)$ is a consistent subsignature and it is also maximal. It follows $\mcsig^\boxminus(K)=\{\atoms(K)\}$. From $\mcsig^\boxminus(K)=\{\atoms(K)\}$ and property 3 of Definition~\ref{def:forget} it follows that $K|^\boxminus_\atoms(K)=K\boxminus \atoms(K)\setminus\atoms(K)=K\boxminus\emptyset=K$ is consistent, showing the equivalence of all three statements.
				\item This follows directly from item 1 of Corollary~\ref{cor:varelem:vs:naiveforget:1}.
				\item This follows directly from item 2 of Corollary~\ref{cor:varelem:vs:naiveforget:1}.\qedhere
			\end{enumerate}
		\end{proof}
	\end{proposition}
Note that item 1 of the above proposition is true for any forgetting operator that complies with Definition~\ref{def:forget}.

A particular property of the set of all minimal inconsistent subsets $\mi(K)$ is its monotony wrt.\ expansions of $K$. More precisely, if $K\subseteq K'$ then $\mi(K)\subseteq \mi(K')$. For the corresponding semantical counterpart $\misig^\boxminus(K)$, this is not generally true.
	\begin{example}
		Consider
		\begin{align*}
			K_4 & = \{a\vee b, \neg a \wedge \neg b\}
		\end{align*}
		Here we have $\misig^{\varelem}(K_4)=\{\{a,b\}\}$. However, adding the formula $a$ gives us
		\begin{align*}
			\misig^{\varelem}(K_4\cup\{ a\})	 & = \{\{a\}\}
		\end{align*}
		and therefore $\misig^{\varelem}(K_4)\not\subseteq \misig^{\varelem}(K_4\cup\{\neg a\})$.
	\end{example}
	But $\misig^\boxminus(K)$ behaves monotonically when it comes to expansions of the signature.
	\begin{proposition}\label{prop:mi:mono}
		Let $K$ be a knowledge base, $S\subseteq\atoms(K)$, and $\boxminus$ a forgetting operator. Then
		\begin{align*}
			\misig^\boxminus(K\boxminus S) & \subseteq \misig^\boxminus(K)
		\end{align*}
		\begin{proof}
			Let $S'\in\misig^\boxminus(K\boxminus S)$. Then for $S''=\atoms(K\boxminus S)\setminus S'$ we have
			\begin{enumerate}
				\item $(K\boxminus S)\boxminus S''\models \perp$ and
				\item for all $S'''\supsetneq S''$, $(K\boxminus S)\boxminus S'''\not\models \perp$.
			\end{enumerate}
			Due to item 4 of Definition~\ref{def:forget} the above is equivalent to
			\begin{enumerate}
				\item $(K\boxminus (S\cup S''))\models \perp$ and
				\item for all $S'''\supsetneq S''$, $(K\boxminus (S\cup  S'''))\not\models \perp$.
			\end{enumerate}
			It follows that $\atoms(K)\setminus(S\cup S'')$ is a minimal inconsistent subsignature of $K$ and
			\begin{align*}
				\atoms(K)\setminus(S\cup S'') & = \atoms(K)\setminus(S\cup (\atoms(K\boxminus S)\setminus S'))	\\
				& =	 \atoms(K)\setminus(\atoms(K)\setminus S'))	\\
				& = S'
			\end{align*}
			and therefore $S'\in\misig^\boxminus(K)$.
		\end{proof}
	\end{proposition}
Another particularly interesting property of $\mi(K)$ and $\mc(K)$ is the \emph{hitting set duality} \cite{Reiter:1987}. A \emph{hitting set} of a set of sets $M=\{M_1,\ldots,M_n\}$ is a set $H\subseteq M_1\cup\ldots\cup M_n$ such that $H\cap M_i\neq \emptyset$ for all $i=1,\ldots,n$. A hitting set $H$ is \emph{minimal} if there is no other hitting set $H'$ with $H'\subsetneq H$. The hitting set duality for $\mi(K)$ and $\mc(K)$ says that $H$ is a minimal hitting set of $\mi(K)$ iff $K\setminus H\in \mc(K)$. Interestingly, we obtain the same duality for $\misig^\boxminus(K)$ and $\mcsig^\boxminus(K)$ and wrt.\ any forgetting operator that complies with Definition~\ref{def:forget}.
	\begin{proposition}\label{prop:hitting}
		Let $\boxminus$ be a forgetting operator and $K$ a knowledge base.
		$H$ is a minimal hitting set of $\misig^\boxminus(K)$ iff $\atoms(K)\setminus H\in \mcsig^\boxminus(K)$.
		\begin{proof}
			Let $\boxminus$ be a forgetting operator.
			\begin{description}
				\item[``$\Rightarrow$'':] Let $H$ be a minimal hitting set of $\misig^\boxminus(K)$ and define $S=\atoms(K)\setminus H$. Assume that $S$ is an inconsistent subsignature. Then there is $M\subseteq S$ and $M$ is a minimal inconsistent subsignature of $K|^{\boxminus}_S=K\boxminus H$. Due to Proposition~\ref{prop:mi:mono} it follows that $M$ is also a minimal inconsistent subsignature of $K$. Then $H$ could not have been a minimal hitting set of $\misig^\boxminus(K)$. It follows that $S$ is a consistent subsignature. Assume there is $a\in\atoms(K)$ such that $S\cup\{a\}$ is also a consistent subsignature. As $H$ is a minimal hitting set, there must be $M\in\misig^\boxminus(K)$ with $M\subseteq S\cup\{a\}$ (otherwise $H\setminus\{a\}$ would also be hitting set and $H$ cannot be minimal). Then $M$ is also a minimal inconsistent subsignature of $K|^{\boxminus}_{S\cup \{a\}}$ and $S\cup\{a\}$ cannot be a consistent subsignature.
				\item[``$\Leftarrow$'':] Let $S\in \mcsig^\boxminus(K)$ and define $H=\atoms(K)\setminus S$. Assume that $H$ is not a hitting set of $\misig^\boxminus(K)$. Then there is $M\in\misig^\boxminus(K)$ and $M\subseteq S$. It follows that $S$ is not a consistent subsignature. Assume that $H$ is not a minimal hitting set, so there is $a\in\atoms(K)$ such that $H\setminus\{a\}$ is also a hitting set. It follows analogously that $S\cup\{a\}$ must be a consistent subsignature, contradicting the fact that $S$ is a maximal consistent subsignature. It follows that $H$ is a minimal hitting set. 	\qedhere
			\end{description}
		\end{proof}
	\end{proposition}
A corollary of the above result is that free propositions can also be characterised as those propositions that appear in all maximal consistent subsignatures (as it is the case with free formulas and maximal consistent subsets).
\begin{corollary}
	Let $\boxminus$ be a forgetting operator and $K$ a knowledge base. A proposition $a\in\atoms(K)$ is a free proposition in $K$ wrt.\ $\boxminus$ iff $a\in S$ for all $S\in \mcsig^\boxminus(K)$.
	\begin{proof}
		Let $a\in\atoms(K)$ be a free proposition of $K$, i.\,e., $a\notin S$ for all $S\in\misig^\boxminus(K)$. It follows that $a\notin H$ for every minimal hitting set $H$ of $\misig^\boxminus(K)$. Due to Proposition~\ref{prop:hitting}, every maximal consistent subsignature $S'\in\mcsig^\boxminus(K)$ has the form $S'\atoms(K)\setminus H$ for some minimal hitting set $H$ of $\misig^\boxminus(K)$. Since $a\notin H$, it follows $h\in S'$.
	\end{proof}
\end{corollary}
\section{Reasoning with maximal consistent subsignatures}\label{sec:maxcons}
As mentioned before, reasoning with maximal consistent subsets of a knowledge bases is a classical approach to reasoning with inconsistent knowledge bases \cite{Rescher:1970,Konieczny:2019}. Using our framework of maximal consistent subsignatures, we can define an analogous approach as follows.
\begin{definition}
	Let $\boxminus$ be a forgetting operation and $K$ a knowledge base. Define the set $\mckb^\boxminus(K)$ via
	\begin{align*}
		\mckb^\boxminus(K) &= \{K|^\boxminus_S\mid S\in \mcsig^\boxminus(K)\}
	\end{align*}
\end{definition}
In other words, $\mckb^\boxminus(K)$ is the set of all knowledge bases projected on maximal consistent subsignatures of $K$.

Some straightforward observations are as follows.
\begin{proposition}\label{prop:misc2}
	Let $\boxminus$ be a forgetting operation and $K$ a knowledge base.
	\begin{enumerate}
		\item $K$ is consistent iff $\mckb^\boxminus(K)=\{K\}$
		\item $\mckb^{\varelem}(K)\neq \emptyset$ iff there is no $\phi\in K$ with $\phi\models\perp$.
		\item $\mckb^{\naiveforget}(K)\neq \emptyset$.
	\end{enumerate}
	\begin{proof}~
	\begin{enumerate}
		\item Assume $K$ is consistent. Due to item~3 of Definition~\ref{def:forget} it follows that $K\boxminus \emptyset = K $ is consistent. Furthermore, $\atoms(K)$ is the only maximal consistent subsignature of $K$ and therefore $K|^\boxminus_{\atoms(K)}=K\boxminus(\atoms(K)\setminus\atoms(K)=K\boxminus\emptyset = K $ and therefore $MCKB^\boxminus(K)=\{K\}$. The other direction is analogous.
		\item This follows from item 2 of Proposition~\ref{prop:misc}.
		\item This follows from item 3 of Proposition~\ref{prop:misc}.
		\qedhere
	\end{enumerate}
	\end{proof}
\end{proposition}
Following \cite{Rescher:1970}, we can define inference relations $\infer^{\boxminus}_{i}$ (\emph{inevitable consequences}) and $\infer^{\boxminus}_{w}$ (\emph{weak consequences}) via
\begin{align*}
	K \infer^{\boxminus}_{i} \alpha &\text{~~iff~~} K' \models \alpha\text{~for all~} 	K'\in \mckb^\boxminus(K)\\
	K \infer^{\boxminus}_{w} \alpha &\text{~~iff~~} K' \models \alpha\text{~for some~} 	K'\in \mckb^\boxminus(K)
\end{align*}
for all knowledge bases $K$ and formulas $\alpha$.
\begin{example}
	We continue Example~\ref{ex:mimc} and consider
	\begin{align*}
		K_3 & = \{a\wedge b \wedge d, \neg a \vee \neg b, b\wedge \neg c, (c \vee \neg b)\wedge d\}
	\end{align*}
	with $\mcsig^{\naiveforget}(K_3) = \{\{a,c,d\},\{b,d\}\}$.
	We have
	\begin{align*}
		\mckb^{\naiveforget}(K_3)	& = \{\{a\wedge d,\neg c,d\},\{b\wedge d, b, d\}\}
	\end{align*}
	and therefore (for example)
	\begin{align*}
		K &\infer^{\naiveforget}_{i} d\\
		K &\infer^{\naiveforget}_{w} d\\
		K &\infer^{\naiveforget}_{w} a\wedge d \\
		K &\notinfer^{\naiveforget}_{w} a\wedge b
	\end{align*}
\end{example}
\begin{example}
	Consider again
	\begin{align*}
		K_1 & = \{a,\neg a\wedge c\}	 &
		K_2 & = \{a\wedge \neg a, c\}
	\end{align*}
	from the introduction. Here we get
	\begin{align*}
		K_1	& \infer^{\naiveforget}_{i} c	& K_1	&\infer^{\naiveforget}_{w} c\\
		K_2	& \infer^{\naiveforget}_{i} c	& K_2	&\infer^{\naiveforget}_{w} c
	\end{align*}
	as initially desired.
\end{example}
We now make some general observations. First, the new inference relations coincide with classical entailment in case of consistent knowledge bases.
\begin{proposition}\label{prop:classical}
Let $\boxminus$ be a forgetting operation. For consistent $K$, $K\infer^{\boxminus}_{i}\alpha$ iff $K\infer^{\boxminus}_{w}\alpha$ iff $K\models \alpha$.
	\begin{proof}
		Proposition~\ref{prop:misc2} item 1 states that a knowledge base $K$ is consistent iff $\mckb^\boxminus(K)=\{K\}$. Since $K \infer^{\boxminus}_{i} \alpha$ iff $K' \models \alpha$ for all $K'\in \mckb^\boxminus(K)$ we have $K \infer^{\boxminus}_{i} \alpha$ iff $K\models \alpha$ (and the same holds for $\infer^{\boxminus}_{w}$).
	\end{proof}
\end{proposition}
We now investigate the compliance of $\infer^{\boxminus}_{i}$ and $\infer^{\boxminus}_{w}$ with some properties from non-monotonic reasoning \cite{Kraus:1990}.
\begin{theorem}
	Let $\boxminus$ be a forgetting operation, \mbox{$\infer\in\{\infer^{\boxminus}_{i},\infer^{\boxminus}_{w}\}$}, $K,K'$ knowledge bases, and $\alpha,\beta$ formulas. The relation $\infer$ satisfies the following properties:
	\begin{description}
		\item[Reflexivity] For consistent $\alpha$, $\{\alpha\}\infer\alpha$.
		\item[Left Logical Equivalence] For consistent $K$ and $K'$, if $K\equiv K'$ and $K\infer \alpha$ then $K'\infer \alpha$.
		\item[Right Weakening] If $\alpha\models\beta$ and $K\infer \alpha$ then $K\infer \beta$.
		\item[Cut] If $K\cup\{\alpha\}\infer \beta$ and $K\infer \alpha$ then $K\infer \beta$.
		\item[Cautious Monotonicity] If $K\infer \alpha$ and $K\infer \beta$ then $K\cup\{\alpha\}\infer \beta$.
	\end{description}
	\begin{proof}
		We show all statements wrt.\ $\infer^{\boxminus}_{i}$, the case $\infer^{\boxminus}_{w}$ is analogous.
		\begin{description}
		\item[Reflexivity] This follows from Proposition~\ref{prop:classical} and the fact that $\{\alpha\}\models \alpha$.
		\item[Left Logical Equivalence] This follows from Proposition~\ref{prop:classical} and the fact that $K\equiv K'$ and $K\models \alpha$ implies $K'\models \alpha$.
		\item[Right Weakening] If $K\infer^{\boxminus}_{i}\alpha$ then for all $K'\in\mckb^\boxminus(K)$, $K'\models \alpha$. If $\alpha\models\beta$ then for all $K'\in\mckb^\boxminus(K)$, $K'\models \beta$. It follows $K\infer^{\boxminus}_{i}\beta$.
		\item[Cut] Let $K\cup\{\alpha\}\infer^{\boxminus}_{i} \beta$ and $K\infer^{\boxminus}_{i} \alpha$. Due to $K\infer^{\boxminus}_{i} \alpha$ it follows that for all $K'\in\mckb^\boxminus(K)$, $K'\models\alpha$. Without loss of generality, assume now  $\atoms(\alpha)\subseteq \atoms(K')$ (otherwise $\alpha$ is of the form $\alpha'\vee\alpha''$ with $\atoms(\alpha'')\cap \atoms(K')=\emptyset$, then we just consider $\alpha'$). It follows that $K'\cup \{\alpha\}$ is consistent. It follows that $K'\in\mckb^\boxminus(K)$ iff $K'\cup\{\alpha\}\in\mckb^\boxminus(K\cup\{\alpha\})$. As $K\cup\{\alpha\} \infer^{\boxminus}_{i}\beta$, it follows that $K'\cup\{\alpha\}\models\beta$ for all $K'\cup\{\alpha\}\in\mckb^\boxminus(K\cup\{\alpha\})$. As $K'\models \alpha$ it follows $K'\models\beta$ and therefore $K\infer^{\boxminus}_{i} \beta$.
		\item[Cautious Monotonicity] Let $K\infer^{\boxminus}_{i}\alpha$ and $K\infer^{\boxminus}_{i}\beta$. It follows that for all $K'\in\mckb^\boxminus(K)$, $K'\models \alpha$ and $K'\models \beta$. Without loss of generality, we can assume $\atoms(\alpha)\subseteq\atoms(K')$ and $\atoms(\beta)\subseteq\atoms(K')$. Then $K'\cup\{\alpha\}\models\beta$ as well. It also follows that $K'\in\mckb^\boxminus(K)$ iff $K'\cup\{\alpha\}\in\mckb^\boxminus(K\cup\{\alpha\})$ and therefore $K\cup \{\alpha\}\infer^{\boxminus}_{i}\beta$.	\qedhere
	\end{description}
	\end{proof}
\end{theorem}
It has to be noted that $\infer\in\{\infer^{\boxminus}_{i},\infer^{\boxminus}_{w}\}$ does not satisfy the standard unconditional versions of reflexivity and left logical equivalence (without the consistency assumptions), since $\infer$ is not explosive. A deeper analysis of the properties from \cite{Kraus:1990} is left for future work.

We conclude our analysis on the formal properties of our new inference relations with an example highlighting the difference to reasoning with maximal consistent subsets.
\begin{example}
	Consider $K_5$ defined via
	\begin{align*}
		K_5 & = \{a\wedge b, b, \neg b \vee \neg a\}
	\end{align*}
	If we reason via maximal consistent \emph{subsets} we get
	\begin{align*}
		K_5 \infermaxcons_i b,
	\end{align*}
	so $b$ is an inevitable consequence (since $\{a\wedge b,\neg b\vee \neg a\}$ is the only minimal inconsistent subset of $K_5$). This seems odd, since $b$ is a central ``ingredient'' in deriving the inconsistency in $K_5$. For our approach we get (for $\boxminus\in\{\varelem,\naiveforget\}$)
	\begin{align*}
		K_5\notinfer^{\boxminus}_{i} b
	\end{align*}
 	but only $K_5\infer^{\boxminus}_{w} b$ and $K_5\infer^{\boxminus}_{w} a$ (as $\{a,b\}$ is the only minimal inconsistent subsignature).
\end{example}
\section{Computational complexity}\label{sec:complexity}
We assume familiarity with the standard complexity classes \textsf{NP} and \textsf{coNP}, see \cite{Papadimitriou:1994a} for an introduction. We also require knowledge of the complexity class $\textsf{DP}$, which is defined as $\textsf{DP}=\{L_1\cap L_2\mid L_1\in \textsf{NP}, L_2\in\textsf{coNP}\}$, i.\,e., the class of problems that are the intersection of a problem in $\textsf{NP}$ and a problem in \textsf{coNP}. We also use complexity classes of the polynomial hierarchy that can be defined (using oracle machines) via	$\Sigma^\textsf{P}_1  = \textsf{NP}$, $\Pi^\textsf{P}_1 = \textsf{coNP}$ 	and
	\begin{align*}
		\Sigma^\textsf{P}_{i} & = \textsf{NP}^{\Sigma^\textsf{P}_{i-1}} &\Pi^\textsf{P}_i & = \textsf{coNP}^{\Sigma^\textsf{P}_{i-1}}
	\end{align*}
	for all $i>1$, where $\mathcal{C}^{\mathcal{D}}$ denotes the class of decision problems solvable in class $\mathcal{C}$ with access to an oracle that can solve problems that are complete for $\mathcal{D}$.

Complexity results for the concrete forgetting operators $\varelem$ and $\naiveforget$ regarding some basic decision problems are as follows. The proof of the following theorem and the remaining technical results from this section can be found in the supplementary material.
\begin{theorem}\label{th:comp:varelem}
	Let $\boxminus\in\{\varelem,\naiveforget\}$, $K$ be a knowledge base, and $S\subseteq\atoms(K)$.
	\begin{enumerate}
		\item Deciding whether $S$ is a consistent subsignature of $K$ wrt.\ $\boxminus$ is \textsf{NP}-complete.
		\item Deciding whether $S$ is a minimally inconsistent subsignature wrt.\ $\boxminus$ is \textsf{DP}-complete.
		\item Deciding whether $S$ is a maximally consistent subsignature wrt.\ $\boxminus$ is \textsf{DP}-complete.
	\end{enumerate}
	\begin{proof}
		We first consider the case of $\boxminus=\varelem$.
		Let $K$ be a knowledge base and $S=\{a_1,\ldots,a_n\}\subseteq\atoms(K)$. Let $\overline{S}=\atoms(K)\setminus S=\{b_1,\ldots,b_m\}$.
		\begin{enumerate}
			\item Let $\phi\in K$. First, observe that computing $\phi\varelem \overline{S}$ directly leads to an exponential blow-up in the size of $\phi$. More precisely, $\phi\varelem \overline{S}$ is of the form
				\begin{align*}
					\phi\varelem \overline{S} & \equiv \bigvee_{x_1,\ldots,x_m\in\{\top,\perp\}}\phi[b_1\rightarrow x_1]\ldots[b_m\rightarrow x_m]
				\end{align*}
				However, in order to satisfy $\phi\varelem \overline{S}$ by an interpretation $\omega:S\rightarrow\{\true,\false\}$, only one of the disjuncts of the above form has to be satisfied. In order to do that, we require a single assignment of $\top,\perp$ to the propositions $b_1,\ldots,b_m$.
		So, we guess (in a non-deterministic step) for each formula $\phi\in K$ an interpretation $\omega_\phi:\atoms(K)\rightarrow \{\true,\false\}$ such that $\omega_{\phi}(a)=\omega_{\phi'}(a)$ for all $a\in S$ and $\phi,\phi'\in K$ (equivalently, each $\omega_{\phi}$ provides an individual assignment of $\top,\perp$ to the propositions in $\overline{S}$). Then we verify (in polynomial time) whether each $\omega_\phi$ satisfies $\phi$. If this is the case, $S$ is a consistent subsignature. This shows \textsf{NP}-membership.

		For \textsf{NP}-hardness, we reduce the problem of checking whether a propositional formula $\phi$ is consistent to our problem. In fact, observe that this problem is equivalent to checking whether $\atoms(\phi)$ is a consistent subsignature of $\{\phi\}$.

		\item For $\textsf{DP}$-membership, observe that deciding whether $S$ is a minimally inconsistent subsignature has two sub-tasks:
			\begin{enumerate}
				\item Deciding whether $S$ is an inconsistent subsignature and
				\item Deciding whether each $S\setminus\{a\}$ is a consistent subsignature (for each $a\in S$).
			\end{enumerate}
			Problem (a) is \textsf{coNP}-complete (follows directly from 1.). For  problem (b), we can adapt the \textsf{NP}-membership proof of 1  by guessing not one $\omega_\phi$ per formula $\phi$, but $|S|$ different ones, one for each subsignature $S\setminus\{a\}$. This only requires polynomially more effort and the problem stays in $\textsf{NP}$ (and the \textsf{NP}-hardness proof remains the same since checking only one signature is a special case). So (b) is \textsf{NP}-complete, which shows that the overall problem is in \textsf{DP}.

			For \textsf{DP}-hardness, we reduce the problem of checking whether for a pair of formulas $(\phi,\psi)$ it is the case that $\phi$ is consistent and $\psi$ is inconsistent. Let $a$ be a new proposition such that $a\notin \atoms(\phi)\cup\atoms(\psi)$. We claim that $\phi$ is consistent and $\psi$ is inconsistent iff $\{a\}$ is a minimally inconsistent subsignature of
				\begin{align*}
					K_{\phi,\psi} & = \{(\phi \Rightarrow a , \neg \psi \Rightarrow \neg a\}
				\end{align*}
				In order to see this, first assume that $\phi$ is consistent and $\psi$ is inconsistent. Since $\phi$ is consistent, it follows that $\phi\varelem\atoms(\phi)\equiv \top$. Similarly, for inconsistent $\psi$ it follows $\psi\varelem\atoms(\psi)\equiv \perp$. So
				\begin{align*}
					K_{\phi,\psi}|^{\varelem}_{\{a\}} & = \{(\phi \Rightarrow a)\varelem\atoms(\phi) , (\neg \psi \Rightarrow \neg a)\varelem(\atoms(\phi)\}\\
					& \equiv \{\top\Rightarrow a,\neg\perp\Rightarrow\neg a\}\\
					& \equiv \{a,\neg a\}
				\end{align*}
				is clearly inconsistent. Moreover, $\{a\}$ is also a minimally inconsistent subsignature, since its only subset $\emptyset$ is consistent. For the other direction, assume that $\{a\}$ is a minimally inconsistent subsignature of $K_{\phi,\psi}$. Since $\phi$ and $\psi$ do not mention $a$, both projections of $\phi$ and $\psi$ onto $a$ can only be $\top$ or $\perp$, respectively. It follows that
					\begin{itemize}
						\item $K_{\phi,\psi}|^{\varelem}_{\{a\}} \equiv \{\top\Rightarrow a,\neg\top\Rightarrow\neg a\}\}$, or
						\item $K_{\phi,\psi}|^{\varelem}_{\{a\}} \equiv \{\perp\Rightarrow a,\neg\perp\Rightarrow\neg a\}\}$, or
						\item $K_{\phi,\psi}|^{\varelem}_{\{a\}} \equiv \{\perp\Rightarrow a,\neg\top\Rightarrow\neg a\}\}$, or
						\item $K_{\phi,\psi}|^{\varelem}_{\{a\}} \equiv \{\top\Rightarrow a,\neg\perp\Rightarrow\neg a\}\}$.
					\end{itemize}
					From the above, only the last one, $K_{\phi,\psi}|^{\varelem}_{\{a\}} \equiv \{\top\Rightarrow a,\neg\perp\Rightarrow\neg a\}\}$, is inconsistent, and due to $\{a\}$ being an inconsistent subsignature, the necessary result. It follows $\phi\varelem\atoms(\phi)\equiv \top$ and $\psi\varelem\atoms(\psi)\equiv \perp$ and therefore $\phi$ is consistent and $\psi$ is inconsistent.
		\item For $\textsf{DP}$-membership, observe that deciding whether $S$ is a maximally consistent subsignature has two sub-tasks:
			\begin{enumerate}
				\item Deciding whether $S$ is a consistent subsignature and
				\item Deciding that there is no consistent subsignature $S'$ with $S\subsetneq S'\subseteq \atoms(K)$.
			\end{enumerate}
			Problem is (a) \textsf{NP}-complete due to 1. For  problem (b), we show that the complement problem, i.\,e., deciding that there is a consistent subsignature $S'$ with $S\subsetneq S'\subseteq \atoms(K)$, is \textsf{NP}-complete. Then it follows that (b) is \textsf{coNP}-complete. For \textsf{NP}-membership, we slightly extend the non-deterministic algorithm from 1 by first guessing the set $S'$ and then proceed as before. For \textsf{NP}-hardness, observe that consistency of a formula $\phi$ (an \textsf{NP}-hard problem to decide) is equivalent to asking whether there is a consistent subsignature $S'$ with $\atoms(\phi)\setminus\{a\}\subsetneq S\subset \atoms(\phi)$ for any one $a\in \atoms(\phi)$ (it necessarily has to hold $S'=\atoms(\phi)$ in that case). This shows that the original problem is in \textsf{DP}.

			For \textsf{DP}-hardness, we reduce the problem of checking whether for a pair of formulas $(\phi,\psi)$ it is the case that $\phi$ is consistent and $\psi$ is inconsistent. Without loss of generality, we assume $\atoms(\phi)\cap\atoms(\psi)=\emptyset$. Let $a$ be a new proposition such that $a\notin \atoms(\phi)\cup\atoms(\psi)$. We claim that $\phi$ is consistent and $\psi$ is inconsistent iff $\atoms(\phi)\cup\atoms(\psi)$ is a maximally consistent subsignature of
				\begin{align*}
					K_{\phi,\psi} & = \{(\phi \wedge a , \neg \psi \wedge \neg a\}
				\end{align*}
				First, observe that
				\begin{align*}
					K_{\phi,\psi}|^{\varelem}_{\atoms(\phi)\cup\atoms(\psi)}& \equiv \{\phi,\neg\psi\}
				\end{align*}
				If $\phi$ is consistent and $\psi$ is inconsistent, then $K_{\phi,\psi}|^{\varelem}_{\atoms(\phi)\cup\atoms(\psi)}$ is obviously consistent. Moreover, the only strict superset of $\atoms(\phi)\cup\atoms(\psi)$ is $\atoms(\phi)\cup\atoms(\psi)\cup\{a\}$ and due to
				\begin{align*}
					K_{\phi,\psi}|^{\varelem}_{\atoms(\phi)\cup\atoms(\psi)\cup\{a\}}& = K_{\phi,\psi}  = \{(\phi \wedge a , \neg \psi \wedge \neg a\}
				\end{align*}
				this is not a consistent subsignature. It follows that $\atoms(\phi)\cup\atoms(\psi)$ is a maximally consistent subsignature. For the other direction, assume that $\atoms(\phi)\cup\atoms(\psi)$ is a maximally consistent subsignature, so
				\begin{align*}
				K_{\phi,\psi}|^{\varelem}_{\atoms(\phi)\cup\atoms(\psi)}& \equiv \{\phi,\neg\psi\}
				\end{align*}
				must be consistent. This can only be the case of $\phi$ is consistent and $\psi$ is inconsistent, concluding the proof.
		\end{enumerate}
		As for $\naiveforget$, \textsf{NP}-membership resp.\ \textsf{DP}-memberships follow directly from above and the fact that we can represent $\naiveforget$ through $\varelem$ via the (polynomial) transformation from Corollary~\ref{cor:naive:varelem}. \textsf{NP}-hardness resp.\ \textsf{DP}-hardness proofs are identical to the corresponding proofs above (in particular, note that the proposition $a$ introduced for the \textsf{DP}-hardness proofs only appears once in each individual formula).
	\end{proof}
\end{theorem}
Since the computational complexities of basic problems regarding $\varelem$ and $\naiveforget$ are the same, we can also state our next results in a more general fashion.
\begin{lemma}\label{lem:infer:comp}
	Let $\boxminus\in\{\varelem,\naiveforget\}$. For given $K$, $\alpha$, and $S\subseteq\atoms(K)$, the problem of deciding $K\boxminus S\models \alpha$ is \textsf{coNP}-complete.
	\begin{proof}
		Note first that determining $K\boxminus S$ explicitly is exponential in the size of $K$, so the result is not directly implied by the complexity of general entailment. However, we will use the same technique as already employed in the proof of Theorem~\ref{th:comp:varelem}, item 1.	We will consider the complement problem, i.\,e., the problem of deciding that $K\boxminus S\not\models \alpha$, and show that this problem is \textsf{NP}-complete.

		For \textsf{NP}-membership, consider the following non-deterministic algorithm on input $K=\{\phi_1,\ldots,\phi_n\}$, $\alpha$, and $S=\{a_1,\ldots,a_m\}$. Let $\overline{S}=\atoms(K)\setminus S = \{b_1,\ldots,b_l\}$. In a first step, we guess, for each $i=1,\ldots,n$ an assignment $f_i:S\rightarrow\{\top,\perp\}$ and an interpretation $\omega:\overline{S}\rightarrow \{\true,\false\}$. Then we simplify $K$ to
			\begin{align*}
				K' & = \{\phi_1[a_1\rightarrow f_1(a_1)]\ldots[a_m\rightarrow f_1(a_m)],\\
					& \qquad\ldots,\\
					& \qquad \phi_n[a_1\rightarrow f_n(a_1)]\ldots[a_m\rightarrow f_n(a_m)]\}
			\end{align*}
			Note that this can now be done in polynomial time. Finally, we verify that $\omega\models K$ and $\omega\not\models\alpha$.

			For \textsf{NP}-hardness, observe that the \textsf{NP}-complete problem of deciding $K\not\models \alpha$ (which is equivalent to deciding that $K\cup \{\neg \alpha\}$ is satisfiable) can be equivalently written as $K\boxminus\emptyset\not\models \alpha$ and therefore directly reduced to our problem.
	\end{proof}
\end{lemma}
\begin{theorem}
	Let $\boxminus\in\{\varelem,\naiveforget\}$. For given $K$ and $\alpha$,
	\begin{enumerate}
		\item the problem of deciding $K\infer^{\boxminus}_w\alpha$ is $\Sigma^\textsf{P}_2$-complete and
		\item the problem of deciding $K\infer^{\boxminus}_i\alpha$ is $\Pi^\textsf{P}_2$-complete.
	\end{enumerate}
	\begin{proof}~
		\begin{enumerate}
		\item For $\Sigma^\textsf{P}_2$-membership, consider the following non-deterministic algorithm. On input $K=\{\phi_1,\ldots,\phi_n\}$ and $\alpha$ we first guess a set $S=\{a_1,\ldots,a_m\}\subseteq\atoms(K)$ and verify that $S$ is a maximal consistent subsignature of $K$ (since that problem is \textsf{DP}-complete for $\boxminus\in\{\varelem,\naiveforget\}$, see Theorem~\ref{th:comp:varelem}, we can solve it with two calls to an \textsf{NP}-oracle). then we can decide $K\boxminus S\models \alpha$ with another call to an \textsf{NP}-oracle due to Lemma~\ref{lem:infer:comp}. This proves $K\infer^{\boxminus}_w\alpha$.

			For $\Sigma^\textsf{P}_2$-hardness, we consider a reduction from the problem of deciding whether a QBF $\Phi$ of the form
				\begin{align*}
					\Phi = \exists x_1,\ldots,x_n\forall y_1,\ldots,y_m: \phi
				\end{align*}
				with a propositional formula $\phi$ over the variables $x_1,\ldots,x_n,y_1,\ldots,y_m$ evaluates to \emph{true}. We claim that $\Phi$ evaluates to true if and only if for (let $p_i,q_i$ be fresh variables for $i=1,\ldots,n$)
				\begin{align*}
					K_\Phi & = \{p_i,q_i,\neg p_i\vee \neg q_i,\neg p_i\vee x_i,\neg q_i\vee \neg x_i\mid \\
						& \hspace{5cm} i=1,\ldots,n \}
				\end{align*}
				it holds $K_\Phi\infer^{\boxminus}_w\phi$.
				We start with making some general observations on $K_\Phi$. Note that $K_\Phi$ is inconsistent (due to, e.\,g., $p_1,q_1,\neg p_1\vee \neg q_1\in K_\Phi$) and that each $\{p_i,q_i\}$, for $i=1,\ldots,n$, is a minimal inconsistent subsignature (for both $\varelem$ and $\naiveforget$). It follows that every maximal consistent subsignature $S$ of $K_\Phi$ has the form
				\begin{align*}
					S & = \{x_1,\ldots,x_n,s_1,\ldots,s_n\mid\\
						& \hspace{2cm} s_i\in \{p_i,q_i\}	, i\in\{1,\ldots,n\}\}
				\end{align*}
				Let $I_S\subseteq \{1,\ldots,n\}$ be those indices where $p_i\in S$, i.\,e.,
				\begin{align*}
					I_S & = 	\{i\mid p_i\in S\}
				\end{align*}
				It follows that every $K'\in\mckb^\boxminus(K)$ is of the following form (let $S$ be the maximal consistent subset mit $K'=K|^\boxminus_S$):
				\begin{align*}
					K' & = 	\{p_i, \neg p_i \vee x_i\mid i \in I_S\}\cup\\
						& \qquad \{q_i, \neg q_i \vee \neg x_i\mid i \in \{1,\ldots,n\}\setminus I_S\}
				\end{align*}
				In particular, for each $I\subseteq\{1,\ldots,n\}$ there is $K_I\in\mckb^\boxminus(K)$ with
				\begin{align*}
					K_I &\models x_i	\quad \text{for all~}i\in I\\
					K_I &\models \neg x_i	\quad \text{for all~}i\in \{1,\ldots,n\}\setminus I
				\end{align*}
				and it is $\mckb^\boxminus(K)=\{K_I\mid I\subseteq\{1,\ldots,n\}\}$.
			We will now show that $\Phi$ evaluates to true if and only if $K_\Phi\infer^{\boxminus}_w\phi$.
			\begin{itemize}
				\item ``$\Rightarrow$'':	 Assume $\Phi$ is true. So there is an interpretation $\omega_1:\{x_1,\ldots,x_n\}\rightarrow\{\true,\false\}$, such that for all interpretations $\omega_2:\{y_1,\ldots,y_m\}\rightarrow\{\true,\false\}$ it holds $\omega_1,\omega_2\models \phi$. For $I=\{i\mid \omega_1(x_i)=\true\}$ it follows $K_I,\omega_2\models \phi$, for all $\omega_2$. It follows $K_I\models \phi$ and therefore $K_\Phi\infer^{\boxminus}_w\phi$.
				\item ``$\Leftarrow$'': 	Assume $K_\Phi\infer^{\boxminus}_w\phi$. Then there is $I\subseteq \{1,\ldots,n\}$ with $K_I\models \phi$. Define $\omega_1:\{x_1,\ldots,x_n\}\rightarrow\{\true,\false\}$ via $\omega_1(x_i)=\true$ iff $i\in I$ (and $\omega_1(x_i)=\false$ otherwise). Then $\omega_1\models \phi$ and therefore for all $\omega_2:\{y_1,\ldots,y_m\}\rightarrow\{\true,\false\}$, $\omega_1,\omega_2\models \phi$. It follows that $\Phi$ evaluates to true.
			\end{itemize}
		\item Regarding $\Pi^\textsf{P}_2$-membership, we show that the complement problem, i.\,e., the problem of deciding $K\notinfer^{\boxminus}_i\alpha$ is in $\Sigma^\textsf{P}_2$. For that consider the following non-deterministic algorithm. On input $K=\{\phi_1,\ldots,\phi_n\}$ and $\alpha$ we first guess a set $S=\{a_1,\ldots,a_m\}\subseteq\atoms(K)$ and verify that $S$ is a maximal consistent subsignature of $K$ (since that problem is \textsf{DP}-complete for $\boxminus\in\{\varelem,\naiveforget\}$, see Theorem~\ref{th:comp:varelem}, we can solve it with two calls to an \textsf{NP}-oracle). then we can decide $K\boxminus S\not\models \alpha$ with another call to an \textsf{NP}-oracle due to Lemma~\ref{lem:infer:comp}. This proves $K\notinfer^{\boxminus}_i\alpha$.

			For $\Pi^\textsf{P}_2$-hardness, we consider a reduction from the problem of deciding whether a QBF $\Phi$ of the form
				\begin{align*}
					\Phi = \exists x_1,\ldots,x_n\forall y_1,\ldots,y_m: \phi
				\end{align*}
				with a propositional formula $\phi$ over the variables $x_1,\ldots,x_n,y_1,\ldots,y_m$ evaluates to \emph{false}. We use the same reduction as in item 1 with a slightly different statement. We claim that $\Phi$ evaluates to false if and only if for
				\begin{align*}
					K_\Phi & = \{p_i,q_i,\neg p_i\vee \neg q_i,\neg p_i\vee x_i,\neg q_i\vee \neg x_i\mid \\
						& \hspace{5cm} i=1,\ldots,n \}
				\end{align*}
				it holds $K_\Phi\infer^{\boxminus}_i\neg\phi$.
			\begin{itemize}
				\item ``$\Leftarrow$'':	We assume $\Phi$ is true. So there is an interpretation $\omega_1:\{x_1,\ldots,x_n\}\rightarrow\{\true,\false\}$, such that for all interpretations $\omega_2:\{y_1,\ldots,y_m\}\rightarrow\{\true,\false\}$ it holds $\omega_1,\omega_2\models \phi$. For $I=\{i\mid \omega_1(x_i)=\true\}$ it follows $K_I,\omega_2\models \phi$, for all $\omega_2$. It follows $K_I\models \phi$ and therefore $K_\Phi\notinfer^{\boxminus}_i\neg\phi$.
				\item ``$\Rightarrow$'': Assume $K_\Phi\notinfer^{\boxminus}_i\neg\phi$. Then there is $I\subseteq \{1,\ldots,n\}$ with $K_I\not\models \neg\phi$. Define $\omega_1:\{x_1,\ldots,x_n\}\rightarrow\{\true,\false\}$ via $\omega_1(x_i)=\true$ iff $i\in I$ (and $\omega_1(x_i)=\false$ otherwise). Then $\omega_1\not\models \neg\phi$ and therefore for all $\omega_2:\{y_1,\ldots,y_m\}\rightarrow\{\true,\false\}$, $\omega_1,\omega_2\not\models \neg\phi$, which is equivalent to $\omega_1,\omega_2\models \phi$. It follows that $\Phi$ evaluates to true.	\qedhere
			\end{itemize}
	\end{enumerate}
	\end{proof}
\end{theorem}

\section{Application to inconsistency measurement}\label{sec:im}
An inconsistency measure \cite{Grant:1978,Thimm:2019d} is a quantitative means to assess the severity inconsistencies in knowledge bases. Let $\mathbb{R}^{\geq 0}_{\infty}$ denote the set of non-negative real numbers including infinity.
\begin{definition}
	An inconsistency measure $I$ is any function $I:2^{\langProp{\atoms}}\rightarrow \mathbb{R}^{\geq 0}_{\infty}$ with $I(K)=0$ iff $K$ is consistent.
\end{definition}
Many existing inconsistency measures are based on minimal inconsistent and maximal consistent subsets of $K$, see \cite{Thimm:2018} for a survey.
We here consider the \textsf{MI}-measure and the \textsf{MI$^C$}-measure by Hunter and Konieczny \cite{Hunter:2008}, as well as the \textsf{MC}-measure and the \emph{problematic measure} by Grant and Hunter \cite{Grant:2011}.
 We can use minimal inconsistent and maximal consistent subsignatures in a similar manner as those measures (we omit the definitions of the original measures here, but their structure is identical to the new measures below). For that, let $\scsig^\boxminus(K)\subseteq \misig^\boxminus(K)$  be the set of all $M\in \misig^\boxminus(K)$ with $|M|=1$ (\emph{self-contradicting} propositions).
\begin{definition}
	Let $\boxminus$ be a forgetting operation and $K$ be a knowledge base. Define functions $I^\boxminus_\text{MISig}$, $I^\boxminus_\text{MISig-C}$, $I^\boxminus_\text{MCSig}$ and $I^\boxminus_\text{P}$ via
	\begin{align*}
		I^\boxminus_\text{MISig}(K) & = \left|\misig^\boxminus(K)\right|\\
		I^\boxminus_\text{MISig-C}(K) & = \sum_{M\in\misig^\boxminus(K)}\frac{1}{|M|}\\
		I^\boxminus_\text{MCSig}(K) & = \left|\mcsig^\boxminus(K)\right|+\left|\scsig^\boxminus(K)\right| -1\\
		I^\boxminus_\text{P}(K) & = \left|\bigcup_{M\in\misig^\boxminus(K)}M\right|
	\end{align*}
\end{definition}
\begin{example}
	We consider again $K_3$ from Example~\ref{ex:mimc} with
		\begin{align*}
			K_3 & = \{a\wedge b \wedge d, \neg a \vee \neg b, b\wedge \neg c, (c \vee \neg b)\wedge d\}
		\end{align*}
		and
		\begin{align*}
			\misig^{\naiveforget}(K_3) &= \{\{a,b\},\{b,c\}\}\\
			\mcsig^{\naiveforget}(K_3) &= \{\{a,c,d\},\{b,d\}\}\\
			\scsig^{\naiveforget}(K_3) &= \emptyset
		\end{align*}
		Here we get
		\begin{align*}
			I^{\naiveforget}_\text{MISig}(K_3) & = 2&
			I^{\naiveforget}_\text{MISig-C}(K_3) & = 1\\
			I^{\naiveforget}_\text{MCSig}(K_3) & = 1&
			I^{\naiveforget}_\text{P}(K_3) & = 3
		\end{align*}
\end{example}
\begin{proposition}
	The functions $I_\text{MISig}$, $I_\text{MISig-C}$, $I_\text{MCSig}$ and $I_\text{P}$ are inconsistency measures.
	\begin{proof}
		We have to show that $I(K)=0$ iff $K$ is consistent for $I\in \{I_\text{MISig}, I_\text{MISig-C}, I_\text{MCSig}, I_\text{P}\}$.
		\begin{itemize}
			\item $I_\text{MISig}$: Due to item 1 of Proposition~\ref{prop:misc} we have that $K$ ist consistent iff $\misig^\boxminus(K)=\emptyset$ iff $I_\text{MISig}=0$.
			\item $I_\text{MISig-C}$: Due to item 1 of Proposition~\ref{prop:misc} we have that $K$ ist consistent iff $\misig^\boxminus(K)=\emptyset$ and we have $I_\text{MISig-C}=\sum\emptyset$ which is defined to be $0$.
			\item $I_\text{MCSig}$: Due to item 1 of Proposition~\ref{prop:misc} we have that $K$ ist consistent iff $\mcsig^\boxminus(K)=\{\atoms(K)\}$ iff $I^\boxminus_\text{MCSig}(K)=0$ (note that $\scsig^\boxminus(K)=\emptyset$ for consistent $K$).
			\item $I^\boxminus_\text{P}$: Due to item 1 of Proposition~\ref{prop:misc} we have that $K$ ist consistent iff $\misig^\boxminus(K)=\emptyset$ and therefore $I^\boxminus_\text{P}=0$.\qedhere
		\end{itemize}
	\end{proof}
\end{proposition}
We leave a further investigation of the above measures for future work.

\section{Relation to paraconsistent reasoning}\label{sec:priest}
We briefly recall Priest's 3-valued logic for paraconsistent reasoning \cite{Priest:1979}.
A \emph{three-valued interpretation} $\upsilon$ on $\atoms$ is a function $\upsilon:\atoms\rightarrow\{T,F,B\}$ where the values $T$ and $F$ correspond to the classical $\true$ and $\false$, respectively. The additional truth value $B$ stands for \emph{both} and is meant to represent a conflicting truth value for a proposition. The function $\upsilon$ is extended to arbitrary formulas as shown in Table~\ref{tbl:para}.
\begin{table}
	\begin{center}
	\begin{tabular}{cc|c|ccc|c}
		$\alpha$ 	& $\beta$ 	& $\upsilon(\alpha\wedge\beta)$ 	&$\upsilon(\alpha\vee\beta)$ 	& \hspace*{0.7cm} 	&	$\alpha$	&	$\upsilon(\neg\alpha)$\\\cline{1-4}\cline{6-7}
		T		   	& 	T		&	T					& T					&					&	T			&	F	\\
		T			&	B		&	B					& T					&					&	B			&	B	\\
		T			&	F		&	F					& T					&					&	F			&	T	\\
		B			&	T		& 	B					& T\\
		B			&	B		&	B					& B\\
		B			&	F		&	F					& B\\
		F			&	T		&	F					& T\\
		F			&	B		&	F					& B\\
		F			&	F		&	F					& F
	\end{tabular}
	\caption{Truth tables for propositional three-valued logic.}
	\label{tbl:para}
	\end{center}
\end{table}
An interpretation $\upsilon$ satisfies a formula $\alpha$ (or is a 3-valued model of that formula), denoted by $\upsilon\models^{3}\alpha$ if either $\upsilon(\alpha)=T$ or $\upsilon(\alpha)=B$. Define $\upsilon\models^{3}K$ for a knowledge base $K$ accordingly. Let $\modelSetP{K}$ denote the set of all 3-valued models of $K$. Note that the interpretation $\upsilon_0$ defined via $\upsilon_0(a)=B$ for all $a\in\atoms$ is a model of every formula, so it makes sense to consider \emph{minimal} models wrt.\ the usage of the paraconsistent truth value $B$. A model $\upsilon$ of a knowledge base $K$ is a \emph{minimal model} of $K$ if it is a model and there is no other model $\upsilon'$ of $K$ with $(\upsilon')^{-1}(B)\subsetneq (\upsilon)^{-1}(B)$. Let $\minModelSetP{K}$ denote the set of minimal models of $K$.

We can define an inference relation on $\minModelSetP{K}$ by considering all minimal models. More formally, define $\infer^3$ via
\begin{align*}
	K \infer^{3} \alpha &\text{~~iff~~} \upsilon \models^3 \alpha\text{~for all~} 	\upsilon\in \minModelSetP{K}
\end{align*}
Note that it makes no sense to define an inference relation by considering only some minimal models, i.\,e., defining $K \infer^{3}_w \alpha$ iff  $\upsilon \models^3 \alpha$ for some $	\upsilon\in \minModelSetP{K}$, since this would give unintuitive results such as $\{a\vee b\}\infer^3_w a\wedge b$.\footnote{This is due to the fact that $\upsilon$ with $\upsilon(a)=\upsilon(b)=T$ is a minimal model of $\{a\vee b\}$ and $\upsilon\models^3 a\wedge b$.}

For a three-valued interpretation $\upsilon$ define its \emph{two-valued} projection $\omega_{\upsilon}:\upsilon^{-1}(\{T,F\})\rightarrow \{\true,\false\}$ via $\omega_{\upsilon}(a)=\true$ iff $\upsilon(a)=T$ and $\omega_{\upsilon}(a)=\false$ iff $\upsilon(a)=F$, for all $a\in \upsilon^{-1}(\{T,F\}$. In other words, $\omega_{\upsilon}$ is a two-valued interpretation that is only defined on those propositions, where $\upsilon$ gives a classical truth value, and the truth value assigned by $\omega_{\upsilon}$ agrees with $\upsilon$.

The relation of three-valued interpretations and forgetting via $\naiveforget$ is characterised as follows. The proof of the following lemma and the remaining technical results from this section can be found in the supplementary material.
\begin{lemma}\label{lem:para1}
	Let $\upsilon$ be a three-valued interpretation. Then
	\begin{align*}
		\upsilon \models^3 \alpha \text{~iff~} \omega_{\upsilon} \models (\alpha\naiveforget\upsilon^{-1}(B))
	\end{align*}
	for every formula $\alpha$.
	\begin{proof}
		We assume that $\alpha$ is in conjunctive normal form\footnote{Observe that $\models^3$ is invariant under the rules of \emph{double negation elimination} ($\neg \neg \alpha\rightarrow \alpha$), \emph{distributivity} ($\alpha \vee (\beta \wedge \gamma) \rightarrow (\alpha\vee \beta)\wedge (\alpha\vee \gamma)$), and the \emph{DeMorgan rules} ($\neg(\alpha\vee \beta)\rightarrow \neg\alpha \wedge \neg\beta$ and $\neg(\alpha\wedge \beta)\rightarrow \neg\alpha \vee \neg\beta$), and therefore invariant under transformation to conjunctive normal form (which amounts to an exhaustive application of these rules), i.\,e., for every $\upsilon$, $\upsilon \models^3 \alpha$ iff $\upsilon \models^3 \alpha'$ when $\alpha'$ is the conjunctive normal form of $\alpha$.}. So let
		\begin{align*}
			\alpha & = C_1\wedge\ldots \wedge C_n
		\end{align*}
		with
		\begin{align*}
			C_i & = l_i^1\vee\ldots\vee l_i^{n_i}
		\end{align*}
		for some $n_i\in \mathbb{N}$ for all $i=1,\ldots, n$ and each $l_i^j$ is of the form $l_i^j=a$ or $l_i^j=\neg a$ for some $a\in \atoms$. Observe that forgetting any atom $a$ that occurs in a sub-formula $C_i$ ($i=1,\ldots,n$) results in making $C_i$ tautological. It follows for any $\atoms'\subseteq \atoms$ that
		\begin{align*}
			\alpha\naiveforget\atoms' & \equiv \hat{C}_1\wedge\ldots\wedge\hat{C}_n
		\end{align*}
		where
		\begin{align*}
			\hat{C}_i & = \left\{\begin{array}{ll}
									\top & \text{if~}\exists j: l_i^j\in\{a,\neg a\mid a\in\atoms'\} \\
									C_i & \text{otherwise}
								\end{array}\right.
		\end{align*}
		We now show that $\upsilon \models^3 \alpha$ implies $\omega_{\upsilon} \models (\alpha\naiveforget\upsilon^{-1}(B))$ and vice versa.
		\begin{itemize}
			\item ``$\Rightarrow$'': Assume $\upsilon \models^3 \alpha$. First observe that the domain of $\omega_\upsilon$ is exactly the signature of $(\alpha\naiveforget\upsilon^{-1}(B))$, since $\omega_\upsilon$ is defined on all atoms that are assigned either $T$ or $F$ by $\upsilon$ and the remaining atoms are forgotten in $(\alpha\naiveforget\upsilon^{-1}(B))$. From $\upsilon \models^3 \alpha$ it follows that $\upsilon(C_i)\in\{T,B\}$ for all $i=1,\ldots,n$.
				\begin{itemize}
					\item Case ``$\upsilon(C_i)=T$'': then there is $l_i^j$ with $\upsilon(l_i^j)=T$ and therefore $\omega_\upsilon(l_i^j)=T$ as well. It follows $\omega_\upsilon\models C_i$.
					\item Case ``$\upsilon(C_i)=B$'': then there is $l_i^j$ with $\upsilon(l_i^j)=B$ and therefore $\hat{C}_i\equiv \top$. Trivially it follows $\omega_\upsilon\models C_i$.
				\end{itemize}
				In summary, it follows $\omega_{\upsilon} \models (\alpha\naiveforget\upsilon^{-1}(B))$.
			\item ``$\Leftarrow$'': Assume $\omega_{\upsilon} \models (\alpha\naiveforget\upsilon^{-1}(B))$. It follows $\omega_{\upsilon}\models \hat{C}_i$ for all $i=1,\ldots,n$. So for each $i=1,\ldots,n$,
				\begin{itemize}
					\item Case ``$\hat{C}_i\equiv \top$'': Then there is $j=1,\ldots,n_i$ such that $l_i^j=a$ or $l_i^j=\neg a$ for some $a$ with $\upsilon(a)=B$. In any case, $\upsilon(C_i)\in\{B,T\}$.
					\item Case ``$\hat{C}_i=C_i$'': Due to $\omega_\upsilon\models \hat{C}_i$ it follows directly that $\upsilon(C_i)=T$.
				\end{itemize}
				In summary, it follows $\upsilon \models^3 \alpha$.
			\qedhere
		\end{itemize}
	\end{proof}
\end{lemma}
In other words, a three-valued interpretation $\upsilon$ is a model of a formula $\alpha$, if and only if the classical part of $\upsilon$ is a model of the formula obtained by forgetting those propositions assigned to $B$.
\begin{lemma}\label{lem:para2}
	Let $K$ be a knowledge base.
	\begin{enumerate}
		\item If $\upsilon\in\minModelSetP{K}$ then $\upsilon^{-1}(B)\in \misig^{\naiveforget}(K)$.
		\item If $S\in \misig^{\naiveforget}(K)$ then there is $\upsilon\in\minModelSetP{K}$ with $\upsilon^{-1}(B)=S$.
	\end{enumerate}
	\begin{proof}~
		\begin{enumerate}
			\item Assume $\upsilon\in\minModelSetP{K}$. So for all $\alpha\in K$, $\upsilon\models^3\alpha$. By Lemma~\ref{lem:para1} it follows that $\omega_\upsilon\models (\alpha\naiveforget\upsilon^{-1}(B))$ and therefore $\omega_\upsilon \models (K\naiveforget\upsilon^{-1}(B)))$. In particular, $(K\naiveforget\upsilon^{-1}(B)))$ is consistent and therefore $\upsilon^{-1}(B)$ is an inconsistent subsignature of $K$. Assume that $\upsilon^{-1}(B)$ is not a minimal inconsistent subsignature of $K$. Then there is $S\subsetneq\upsilon^{-1}(B)$ such that there is an interpretation $\omega$ with $\omega\models(K\naiveforget S)$. Let $\upsilon'$ be such that $\omega_{\upsilon'}=\omega$. By Lemma~\ref{lem:para1} it follows that $\upsilon'\models^3 K$ and $(\upsilon')^{-1}(B)=S\subsetneq \upsilon^{-1}(B)$. Therefore, $\upsilon$ could not have been a minimal model. It follows that $\upsilon^{-1}(B)\in \misig^{\naiveforget}(K)$.
			\item Let $S\in \misig^{\naiveforget}(K)$. Let $\omega$ be such that $\omega\models (K\naiveforget S)$. Define $\upsilon$ such that $\omega=\omega_\upsilon$. By Lemma~\ref{lem:para1} it follows that $\upsilon\models^3 K$. Showing that $\upsilon$ is also a minimal model is analogous as in 1.\qedhere
		\end{enumerate}
	\end{proof}
\end{lemma}
The above lemmas allows us to equate reasoning with three-valued models and reasoning with maximal consistent subsignatures wrt.\ $\naiveforget$ (using the approach of inevitable consequences).
\begin{theorem}
	$\infer^{3}=\infer^{\naiveforget}_{i}$.
	\begin{proof}
		Let $K$ be a knowledge base and $\alpha$ a formula. We only show that $K\infer^3\alpha$ implies $K\infer^{\naiveforget}_i\alpha$, the other direction is analogous.

		Assume $K\infer^3\alpha$, so for all $\upsilon\in \minModelSetP{K}$, $\upsilon \models^3 \alpha$. For $K\infer^{\naiveforget}_i\alpha$ to hold we have to show that
		\begin{align*}
			 K' \models \alpha\text{~for all~} 	K'\in \mckb^{\naiveforget}(K)
		\end{align*}
		which is equivalent to
		\begin{align*}
			\omega\models \alpha\text{~for all~} \omega\text{~with~}\omega\models K|^{\naiveforget}_S \text{~and~}S\in \misig^{\naiveforget}(K)
		\end{align*}
		Let $S\in \misig^{\naiveforget}(K)$ and $\omega$ with $\omega\models K|^{\naiveforget}_S$ be arbitrary. Due to item 2 of Lemma~\ref{lem:para2} and Lemma~\ref{lem:para1} there is  $\upsilon\in \minModelSetP{K}$ with $\omega=\omega_\upsilon$. Since $\upsilon\models^3\alpha$ it follows $\omega_\upsilon\models \alpha$.
	\end{proof}
\end{theorem}
We observed above that the weak variant $\infer^{3}_w$ of $\infer^{3}$ gives unintuitive results (such as $\{a\vee b\}\infer_w^3 a\wedge b$) and is usually not considered. Since $\infer^{3}=\infer^{\naiveforget}_{i}$ one could wonder whether $\infer^{\naiveforget}_w$ exhibits a similar behaviour or even if $\infer^3_w=\infer^{\naiveforget}_w$. This is not the case since $\infer^{\naiveforget}_w$ coincides with classical entailment in case of consistent knowledge bases (see Proposition~\ref{prop:classical}) and $\{a\vee b\}\not\models a\wedge b$. Therefore, $\infer^{\naiveforget}_w$ is a genuinely new inference relation.

\section{Summary and conclusion}\label{sec:conc}
We considered an approach to inconsistency-tolerant reasoning based on forgetting parts of the signature such that the remaining knowledge base is consistent. In particular, we considered the notions of minimal inconsistent and maximal consistent subsignatures as analogons to minimal inconsistent and maximal consistent subsets. Structurally, minimal inconsistent and maximal consistent subsignatures behave similarly as their subset-based counterparts, in particular, we showed that the hitting set duality is also satisfied by those notions. The induced reasoning relations behave well with respect to a series of desirable properties from non-monotonic reasoning, but are not as syntax-dependent as their subset-based counterparts. Finally, we also investigated relationships to inconsistency measurement and paraconsistent reasoning.

As mentioned before, our framework can be seen as an instance of the more general framework of Lang and Marquis \cite{DBLP:conf/kr/LangM02,Lang:2010}, who broadly investigate the approach of restoring consistency through forgetting. However, the notions of minimal inconsistent and maximal consistent subsignatures are new to our work, as are the induced inference relations. Further related works are by Besnard et al.\ \cite{DBLP:conf/jelia/BesnardSTW02,DBLP:conf/ecsqaru/BesnardSTW03}, who substitute every occurrence of a proposition with a new proposition and use default rules to enforce the identification of these different instances to a maximum degree. Although the technical approach is different to ours, there are some conceptual similarities that we will investigate in more depth for future work.

A further possible venue for future work is to develop signature-based variants of reasoning methods based on maximal consistent subsets other than the classical approach of Rescher and Manor \cite{Rescher:1970}. For example, Konieczny et al.\ \cite{Konieczny:2019} propose inference relations that only consider \emph{some} of the maximal consistent subsets of a knowledge base, where the consideration of maximal consistent subsets is determined by a scoring function. Adapting those scoring functions for maximal consistent subsignatures will therefore give rise to further inference relations. Moreover, the reasoning approach of Brewka \cite{DBLP:conf/ijcai/Brewka89}, who considers \emph{stratified knowledge bases}---i.\,e.\ knowledge bases where formulas are ranked according to their preference---, can also be cast into our framework by considering \emph{stratified signatures}.

\section*{Acknowledgments}
We are grateful to Viorica Sofronie-Stokkermans for discussions.
The research reported here was partially supported by the Deutsche Forschungsgemeinschaft (grant 465447331, project ``Explainable Belief Merging'', EBM).

\bibliographystyle{plain}
\bibliography{references.bib}
\end{document}